\documentclass[runningheads]{llncs}
\let\svthefootnote\thefootnote
\newcommand\blankfootnote[1]{%
  \let\thefootnote\relax\footnotetext{#1}%
  \let\thefootnote\svthefootnote%
}

\usepackage{graphicx}
\usepackage{amsmath,amssymb} 
\usepackage{color, algorithm, algorithmic, subfigure, enumerate, url}

\usepackage[width=122mm,left=12mm,paperwidth=146mm,height=193mm,top=12mm,paperheight=217mm]{geometry}

\newcommand{\enc}{\mathrm{enc}}
\newcommand{\dec}{\mathrm{dec}}
\newcommand{\out}{\mathrm{lab}}

\newcommand{\dis}{c}
\newcommand{\gen}{r}
\newcommand{\convnet}{\textnormal{ConvNet}}
\newcommand{\convnetsrc}{\textnormal{ConvNet}_{src}}
\newcommand{\convnettgt}{\textnormal{ConvNet}_{tgt}}
\newcommand{\sda}{\textnormal{SDA}_{sh}}
\newcommand{\scaet}{\textnormal{SCAE}_t}
\newcommand{\mrcn}{\textnormal{DRCN}}
\newcommand{\bmrcn}{\textbf{DRCN}}
\newcommand{\mrcns}{\textnormal{DRCN}_s}
\newcommand{\bmrcns}{\textbf{DRCN}_s}
\newcommand{\mrcnst}{\textnormal{DRCN}_{st}}
\newcommand{\bmrcnst}{\textbf{DRCN}_{st}}

\renewcommand{\algorithmicrequire}{\textbf{Input:}}
\renewcommand{\algorithmicensure}{\textbf{Output:}}

\newcommand{\cF}{{\mathcal F}}

\newcommand{\cL}{{\mathcal L}}

\newcommand{\cX}{{\mathcal X}}
\newcommand{\cY}{{\mathcal Y}}

\newcommand{\bx}{{\mathbf x}}

\newcommand{\by}{{\mathbf y}}

\newcommand{\bbD}{{\mathbb D}}

\newcommand{\bbP}{{\mathbb P}}
\newcommand{\bbQ}{{\mathbb Q}}
\newcommand{\bbR}{{\mathbb R}}

\DeclareMathOperator*{\argmax}{argmax}
\DeclareMathOperator*{\expec}{\mathbb E}

\begin{document}
\pagestyle{headings}
\mainmatter
\def\ECCV16SubNumber{***}  

\title{Deep Reconstruction-Classification Networks \\ for Unsupervised Domain Adaptation} 

\titlerunning{DRCN for Unsupervised Domain Adaptation}

\authorrunning{M. Ghifary, W. B. Kleijn, M. Zhang, D. Balduzzi, W. Li}

\author{Muhammad Ghifary${}^{1*}$, W. Bastiaan Kleijn${}^1$, Mengjie Zhang${}^1$, David Balduzzi${}^1$, Wen Li${}^2$}
\institute{${}^1$Victoria University of Wellington, ${}^2$ETH Z\"{u}rich}

\maketitle

\blankfootnote{${}^*$ Current affiliation: Weta Digital}

\begin{abstract}
\begin{sloppypar}
In this paper, we propose a novel unsupervised domain adaptation algorithm based on deep learning for visual object recognition.
Specifically, we design a new model called Deep Reconstruction-Classification Network (DRCN), which jointly learns a shared encoding representation for two tasks: 
i) supervised classification of labeled source data, and 
ii) unsupervised reconstruction of unlabeled target data.
In this way, the learnt representation not only preserves discriminability, but also encodes useful information from the target domain. 
Our new DRCN model can be optimized by using backpropagation similarly as the standard neural networks.
\end{sloppypar}

We evaluate the performance of $\mrcn$ on a series of cross-domain object recognition tasks, where $\mrcn$ provides
a considerable improvement (up to $\sim$8$\%$ in accuracy) over the prior state-of-the-art algorithms.
Interestingly, we also observe that the reconstruction pipeline of $\mrcn$ transforms images from the source domain into images whose appearance resembles the target dataset. 
This suggests that $\mrcn$'s performance is due to constructing a single composite representation that encodes information about both the structure of target images and the classification of source images. Finally, we provide a formal analysis to justify the algorithm's objective in domain adaptation context.
\end{abstract}

\section{Introduction}
\label{sec:intro}
An important task in visual object recognition is to design algorithms that are robust to \emph{dataset bias}~\cite{Torralba2011}. 
Dataset bias arises when labeled training instances are available from a source domain and test instances are sampled from a related, but different, target domain.
For example, consider a person identification application in \emph{unmanned aerial vehicles} (UAV), which is essential for a variety of tasks, such as surveillance, people search, and remote monitoring \cite{Hsu:2015}.
One of the critical tasks is to identify people from a bird's-eye view; however collecting labeled data from that viewpoint can be very challenging.
It is more desirable that a UAV can be trained on some already available \emph{on-the-ground} labeled images (source), e.g., people photographs from social media, and then successfully applied to the actual UAV view (target). 
Traditional supervised learning algorithms typically perform poorly in this setting, since they assume that the training and test data are drawn from the same domain.

Domain adaptation attempts to deal with dataset bias using unlabeled data from the target domain so that the task of manual labeling the target data can be reduced.
Unlabeled target data provides auxiliary training information that should help algorithms generalize better on the target domain than using source data only.
Successful domain adaptation algorithms have large practical value, since acquiring a huge amount of labels from the target domain is often expensive or impossible.
Although domain adaptation has gained increasing attention in object recognition, see \cite{patel_dasurvey:2015} for a recent overview, the problem remains essentially unsolved since model accuracy has yet to reach a level that is satisfactory for real-world applications. 
Another issue is that many existing algorithms require optimization procedures that do not scale well as the size of datasets increases \cite{Aljundi_CVPR2015,Baktashmotlagh:2013aa,Bruzzone2010,Gong:2013ab,Long:2013aa,Long2014a,Pan2011}.
Earlier algorithms were typically designed for relatively small datasets, e.g., the Office dataset~\cite{Saenko:2010aa}.

We consider a solution based on learning representations or features from raw data.
Ideally, the learned feature should model the label distribution as well as reduce the discrepancy between the source and target domains.
We hypothesize that a possible way to approximate such a feature is by (supervised) learning the \emph{source label} distribution and (unsupervised) learning of the \emph{target data distribution}.
This is in the same spirit as \emph{multi-task learning} in that learning auxiliary tasks can help the main task be learned better \cite{Caruana1997,Argyriou2006}.
The goal of this paper is to develop an accurate, scalable multi-task feature learning algorithm in the context of domain adaptation.

\paragraph{\textbf{Contribution:}}
To achieve the goal stated above, we propose a new deep learning model for unsupervised domain adaptation. 
Deep learning algorithms are highly scalable since they run in linear time, can handle streaming data, and can be parallelized on GPUs. 
Indeed, deep learning has come to dominate object recognition in recent years~\cite{Krizhevsky_NIPS2012,Simonyan_ICLR2015}. 

We propose \emph{Deep Reconstruction-Classification Network} ($\mrcn$), a convolutional network that jointly learns two tasks: 
i) supervised source label prediction and 
ii) unsupervised target data reconstruction.
The encoding parameters of the $\mrcn$ are shared across both tasks, while the decoding parameters are separated.
The aim is that the learned label prediction function can perform well on classifying images in the target domain -- the data reconstruction can thus be viewed as an auxiliary task to support the adaptation of the label prediction.
Learning in $\mrcn$ alternates between unsupervised and supervised training, which is different from the standard \emph{pretraining-finetuning} strategy \cite{Hinton06afast,Bengio:2007aa}.

From experiments over a variety of cross-domain object recognition tasks, $\mrcn$ performs better than  the state-of-the-art domain adaptation algorithm \cite{Ganin2015}, with up to $\sim8\%$ accuracy gap.
The $\mrcn$ learning strategy also provides a considerable improvement over the pretraining-finetuning strategy, indicating that it is more suitable for the unsupervised domain adaptation setting.
We furthermore perform a visual analysis by reconstructing source images through the learned reconstruction function. 
It is found that \emph{the reconstructed outputs resemble the appearances of the target images} suggesting that the encoding representations are successfully adapted.
Finally, we present a probabilistic analysis to show the relationship between the $\mrcn$'s learning objective and a semi-supervised learning framework \cite{Cohen:2006}, and also the soundness of considering only data from a target domain for the data reconstruction training.

\section{Related Work}
Domain adaptation is a large field of research, with related work under several names such as class imbalance \cite{Japkowicz:2002}, covariate shift \cite{Shimodaira:2000aa}, and sample selection bias \cite{Zadrozny:2004}.
In \cite{Pan:2010aa}, it is considered as a special case of transfer learning.
Earlier work on domain adaptation focused on text document analysis and NLP \cite{Blitzer:2006aa,Daume-III:2007aa}.
In recent years, it has gained a lot of attention in the computer vision community, mainly for object recognition application, see \cite{patel_dasurvey:2015} and references therein. 
The domain adaptation problem is often referred to as \emph{dataset bias} in computer vision \cite{Torralba2011}.

This paper is concerned with \emph{unsupervised domain adaptation} in which labeled data from the target domain is not available \cite{Margolis:2011}.
A range of approaches along this line of research in object recognition have been proposed \cite{Aljundi_CVPR2015,Baktashmotlagh:2013aa,Fernando:2013aa,Ghifary:SCA2015,Gopalan:2011aa,Gong:2012aa,Long2014a}, most were designed specifically for small datasets such as the Office dataset \cite{Saenko:2010aa}.
Furthermore, they usually operated on the SURF-based features \cite{Bay:2008aa} extracted from the raw pixels.
In essence, the unsupervised domain adaptation problem remains open and needs more powerful solutions that are useful for practical situations.

Deep learning now plays a major role in the advancement of domain adaptation.
An early attempt addressed large-scale sentiment classification \cite{Glorot:2011aa}, where the concatenated features from fully connected layers of stacked denoising autoencoders have been found to be domain-adaptive \cite{Vincent:2010aa}.
In visual recognition, a fully connected, shallow network pretrained by denoising autoencoders has shown a certain level of effectiveness \cite{Ghifary2014b}.
It is widely known that deep convolutional networks (ConvNets) \cite{LeCun:1998aa} are a more natural choice for visual recognition tasks and have achieved significant successes \cite{Girshick_CVPR2014,Krizhevsky_NIPS2012,Simonyan_ICLR2015}.
More recently, ConvNets pretrained on a large-scale dataset, ImageNet, have been shown to be reasonably effective for domain adaptation \cite{Krizhevsky_NIPS2012}.
They provide significantly better performances than the SURF-based features on the Office dataset \cite{Donahue:2014aa,Hoffman:2013ab}.
An earlier approach on using a convolutional architecture without pretraining on ImageNet, DLID, has also been explored \cite{Chopra:2013aa} and performs better than the SURF-based features. 

To further improve the domain adaptation performance, the pretrained ConvNets can be \emph{fine-tuned} under a particular constraint related to minimizing a domain discrepancy measure \cite{Ganin2015,Long_DAN:2015,Tzeng_DDC:2014,Tzeng_ICCV2015}.
Deep Domain Confusion (DDC) \cite{Tzeng_DDC:2014} utilizes the maximum mean discrepancy (MMD) measure \cite{Borgwardt:2006aa} as an additional loss function for the fine-tuning to adapt the last fully connected layer.
Deep Adaptation Network (DAN) \cite{Long_DAN:2015} fine-tunes not only the last fully connected layer, but also some convolutional and fully connected layers underneath, and outperforms DDC.
Recently, the deep model proposed in \cite{Tzeng_ICCV2015} extends the idea of DDC by adding a criterion to guarantee the class alignment between different domains.
However, it is limited only to the \emph{semi-supervised} adaptation setting, where a small number of target labels can be acquired.

The algorithm proposed in \cite{Ganin2015}, which we refer to as ReverseGrad, handles the domain invariance as a binary classification problem. It thus optimizes two contradictory objectives: i) minimizing label prediction loss and ii) maximizing domain classification loss via a simple \emph{gradient reversal} strategy.
ReverseGrad can be effectively applied both in the pretrained and randomly initialized deep networks. 
The randomly initialized model is also shown to perform well on cross-domain recognition tasks other than the Office benchmark, i.e., large-scale handwritten digit recognition tasks.
Our work in this paper is in a similar spirit to ReverseGrad in that it does not necessarily require pretrained deep networks to perform well on some tasks. 
However, our proposed method undertakes a fundamentally different learning algorithm: finding a good label classifier while simultaneously learning the structure of the target images.

\section{Deep Reconstruction-Classification Networks}
This section describes our proposed deep learning algorithm for unsupervised domain adaptation, which we refer to as \emph{Deep Reconstruction-Classification Networks} ($\mrcn$).
We first briefly discuss the unsupervised domain adaptation problem. 
We then present the DRCN architecture, learning algorithm, and other useful aspects.

Let us define a \emph{domain} as a probability distribution $\bbD_{XY}$ (or just $\bbD$) on $\cX \times \cY$, where $\cX$ is the input space and $\cY$ is the output space.
Denote the source domain by $\bbP$ and the target domain by $\bbQ$, where $\bbP \neq \bbQ$.
The aim in \emph{unsupervised domain adaptation} is as follows:
given a labeled i.i.d. sample from a source domain $S^s = \{(x^s_i, y^s_i) \}_{i=1}^{n_s} \sim \bbP$ and 
an unlabeled sample from a target domain $S^t_u = \{(x^t_i) \}_{i=1}^{n_t} \sim \bbQ_X$, 
find a good labeling function $f : \cX \rightarrow \cY$ on $S^t_u$.
We consider a feature learning approach: finding a function $g: \cX \rightarrow \cF$ such that the discrepancy between distribution $\bbP$ and $\bbQ$ is minimized in $\cF$.

\begin{figure}[!htb]
  \centering
  \includegraphics[width=4.5in,height=1.8in]{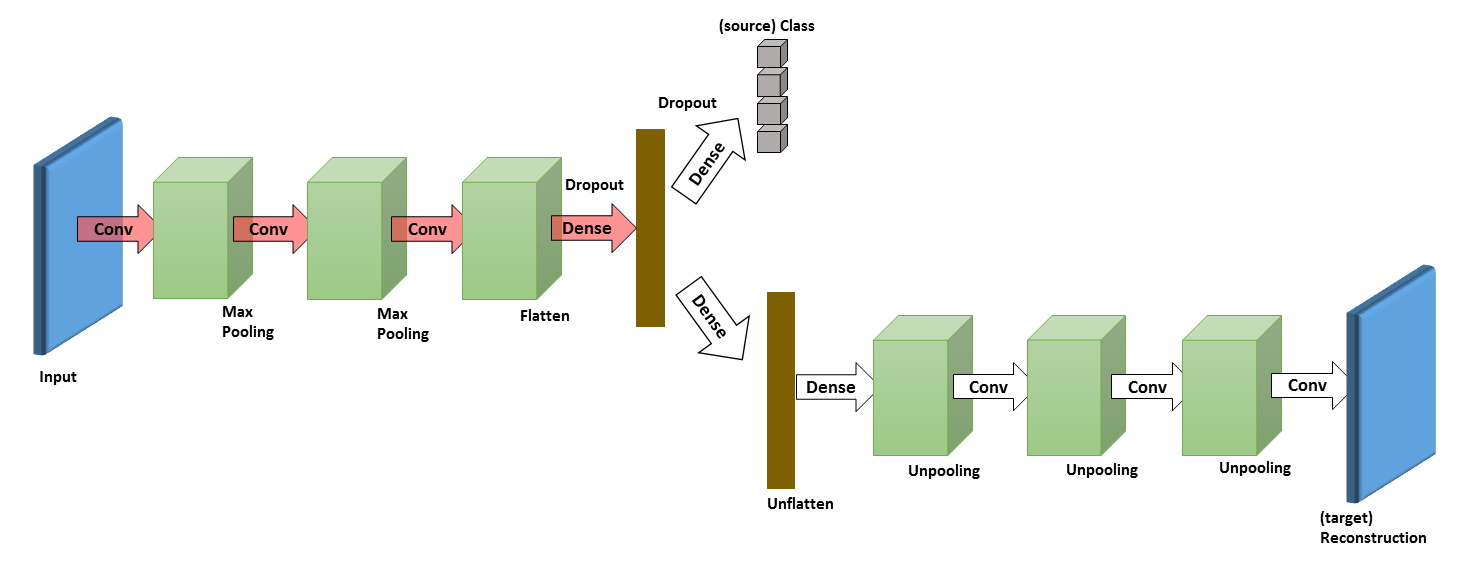}
  \caption{Illustration of the $\mrcn$'s architecture. It consists of two pipelines: i) label prediction and ii) data reconstruction pipelines. The shared parameters between those two pipelines are indicated by the red color.}
  \label{fig:mdgn_arch}
\end{figure}

Ideally, a discriminative representation should model both the label and the structure of the data.
Based on that intuition, we hypothesize that a domain-adaptive representation should satisfy two criteria: 
i) classify well the source domain labeled data and
ii) reconstruct well the target domain unlabeled data, which can be viewed as an approximate of the ideal discriminative representation.
Our model is based on a convolutional architecture that has two pipelines with a shared encoding representation.
The first pipeline is a standard convolutional network for \emph{source label prediction}~\cite{LeCun:1998aa}, while the second one is a convolutional autoencoder for \emph{target data reconstruction}~\cite{Masci2011,Zeiler:2010}.
Convolutional architectures are a natural choice for object recognition to capture spatial correlation of images.
The model is optimized through multitask learning~\cite{Caruana1997}, that is, jointly learns the (supervised) source label prediction and the (unsupervised) target data reconstruction tasks.\footnote{The unsupervised convolutional autoencoder is not trained via the greedy layer-wise fashion, but only with the standard back-propagation over the whole pipeline.}
The aim is that the encoding shared representation should learn the commonality between those tasks that provides useful information for cross-domain object recognition.
Figure~\ref{fig:mdgn_arch} illustrates the architecture of $\mrcn$.

We now describe $\mrcn$ more formally. 
Let $f_\dis : \cX \rightarrow \cY$ be the (supervised) label prediction pipeline and $f_\gen : \cX \rightarrow \cX$  be the (unsupervised) data reconstruction pipeline of $\mrcn$.
Define three additional functions: 
1) an encoder / feature mapping $g_{\enc} : \cX \rightarrow \cF$, 
2) a decoder $g_{\dec} : \cF \rightarrow \cX$, and 
3) a feature labeling $g_{\out}: \cF \rightarrow \cY$.
For $m$-class classification problems, the output of $g_{\out}$ usually forms an $m$-dimensional vector of real values in the range $[0,1]$ that add up to 1, i.e., \emph{softmax} output.
Given an input $x \in \cX$, one can decompose $f_\dis$ and $f_\gen$ such that
\begin{eqnarray}
 \label{eq:mdgn_dpass} f_\dis(x) &=& (g_{\out} \circ g_{\enc} ) (x) , \\
 \label{eq:mdgn_gpass} f_\gen(x) &=& (g_{\dec} \circ g_{\enc} ) (x) .
\end{eqnarray}

Let $\Theta_\dis = \{\Theta_{\enc}, \Theta_{\out} \}$ and $\Theta_\gen = \{\Theta_{\enc}, \Theta_{\dec} \}$ denote the parameters of the supervised and unsupervised model.
$\Theta_{\enc}$ are shared parameters for the feature mapping $g_{\enc}$.
Note that $\Theta_{\enc}, \Theta_{\dec}, \Theta_{\out}$ may encode parameters of multiple layers.
The goal is to seek a single feature mapping $g_{\enc}$ model that supports both $f_\dis$ and $f_\gen$.

\paragraph{\textbf{Learning algorithm:}} The learning objective is as follows.
Suppose the inputs lie in $\cX \subseteq \bbR^d$ and their labels lie in $\cY \subseteq \bbR^m$.
Let $\ell_\dis: \cY \times \cY \rightarrow \bbR$ and $\ell_\gen : \cX \times \cX \rightarrow \bbR$ be the classification and reconstruction loss respectively. 
Given labeled source sample $S^s = \{(\bx^s_i, \by^s_i) \}_{i=1}^{n_s} \sim \bbP$, where $\by_i \in \{ 0, 1\}^m$ is a \emph{one-hot} vector, and unlabeled target sample  $S^t_u = \{(\bx^t_j) \}_{j=1}^{n_t} \sim \bbQ$, we define the empirical losses as:
\begin{eqnarray}
\cL^{n_s}_\dis (  \{ \Theta_{\enc}, \Theta_{\out} \} ) := \sum_{i=1}^{n_s} \ell_\dis \left( f_\dis (\bx^s_i; \{ \Theta_{\enc}, \Theta_{\out} \}), \by^s_i\right), \\
\cL^{n_t}_\gen (  \{ \Theta_{\enc}, \Theta_{\dec} \} ) := \sum_{j=1}^{n_t} \ell_\gen \left( f_\gen (\bx^t_j; \{ \Theta_{\enc}, \Theta_{\dec} \}), \bx^t_j)\right).
\end{eqnarray}
Typically, $\ell_\dis$ is of the form \emph{cross-entropy loss}
$\displaystyle \sum_{k=1}^m y_k \log [f_\dis(\bx)]_k$  (recall that $f_c(\bx)$ is the softmax output)
and $\ell_\gen$ is of the form \emph{squared loss} 
$\displaystyle \| \bx - f_\gen(\bx) \|_2^2$.

Our aim is to solve the following objective:
\begin{equation}
  \label{eq:mdgn_obj}
   \min \lambda \cL^{n_s}_\dis (  \{ \Theta_{\enc}, \Theta_{\out} \} ) + (1-\lambda) \cL^{n_t}_\gen (  \{ \Theta_{\enc}, \Theta_{\dec} \} ),
\end{equation}
where $0 \leq \lambda \leq 1$ is a hyper-parameter controlling the trade-off between classification and reconstruction.
The objective is a convex combination of supervised and unsupervised loss functions. 
We justify the approach in Section \ref{sec:analysis}.

Objective (\ref{eq:mdgn_obj})  can be achieved by alternately minimizing $\cL^{n_s}_\dis$ and $\cL^{n_t}_\gen$ using \emph{stochastic gradient descent} (SGD). 
In the implementation, we used RMSprop~\cite{Tieleman2012}, the variant of SGD with a gradient normalization -- the current gradient is divided by a moving average over the previous root mean squared gradients.
We utilize dropout regularization~\cite{Srivastava_JMLR2014} during $\cL^{n_s}_\dis$ minimization,   which is effective to reduce overfitting.
Note that dropout regularization is applied in the fully-connected/dense layers only, see Figure~\ref{fig:mdgn_arch}.

The stopping criterion for the algorithm is determined by monitoring the average reconstruction loss of the unsupervised model during training -- the process is stopped when the average reconstruction loss stabilizes. 
Once the training is completed, the optimal parameters $\hat{\Theta}_{\enc}$ and $\hat{\Theta}_{\out}$ are used to form a classification model $f_\dis(\bx^t; \{ \hat{\Theta}_{\enc}, \hat{\Theta}_{\out}\})$ that is expected to perform well on the target domain.
The $\mrcn$ learning algorithm is summarized in Algorithm~\ref{alg:mdgn} and implemented using Theano \cite{Theano:2012}.

\begin{algorithm}[!htb]
\footnotesize
	\caption{The Deep Reconstruction-Classification Network ($\mrcn$) learning algorithm.} 
	\label{alg:mdgn}
	\algorithmicrequire\; \\
	$\bullet$ Labeled source data: $S^s = \{( \bx^s_i, y^s_i) \}_{i=1}^{n_s}$; \\
	$\bullet$ Unlabeled target data: $S^t_u = \{ \bx^t_j\}_{i=j}^{n_t}$;\\
	$\bullet$ Learning rates: $\alpha_\dis$ and $\alpha_\gen$;
	\begin{algorithmic}[1]
		\STATE Initialize parameters $\Theta_{\enc}, \Theta_{\dec}, \Theta_{\out}$
		\;
		\WHILE{not stop} 
			\FORALL{source batch of size $m_s$} 
				\STATE{Do a forward pass according to (\ref{eq:mdgn_dpass});}
				\STATE{Let $\Theta_\dis = \{ \Theta_{\enc}, \Theta_{\out} \}$. 
				Update $\Theta_\dis$:}				
				\begin{equation}
					\Theta_\dis \leftarrow \Theta_\dis - \alpha_\dis \lambda \nabla_{\Theta_\dis} \cL^{m_s}_\dis(\Theta_\dis); \nonumber
				\end{equation}
			\ENDFOR
			\FORALL{target batch of size $m_t$} 
				\STATE{Do a forward pass according to (\ref{eq:mdgn_gpass});}
				\STATE{Let $\Theta_\gen = \{ \Theta_{\enc}, \Theta_{\dec} \}$. 
				Update $\Theta_\gen$:}				
				\begin{equation}
					\Theta_\gen \leftarrow \Theta_\gen - \alpha_\gen (1 - \lambda)\nabla_{\Theta_\gen} \cL^{m_t}_\gen(\Theta_\gen). \nonumber
				\end{equation}
			\ENDFOR
		\ENDWHILE
	\end{algorithmic}
\algorithmicensure\\
$\bullet$ $\mrcn$ learnt parameters: $\hat{\Theta} = \{ \hat{\Theta}_{\enc}, \hat{\Theta}_{\dec}, \hat{\Theta}_{\out}\}$;
\end{algorithm}

\paragraph{\textbf{Data augmentation and denoising:}}
We use two well-known strategies to improve $\mrcn$'s performance: data augmentation and denoising.
Data augmentation generates additional training data during the supervised training with respect to some plausible transformations over the original data, which improves generalization, see e.g. \cite{Simard:2003}.
Denoising involves reconstructing \emph{clean} inputs given their \emph{noisy} counterparts. It is used to improve the feature invariance of denoising autoencoders (DAE)~\cite{Vincent:2010aa}.
Generalization and feature invariance are two properties needed to improve domain adaptation.
Since $\mrcn$ has both classification and reconstruction aspects, we can naturally apply these two tricks simultaneously in the training stage.

Let $\bbQ_{\tilde{X} | X}$ denote the noise distribution given the original data from which the noisy data are sampled from.
The classification pipeline of $\mrcn$ $f_\dis$ thus actually observes additional pairs $\{ (\tilde{\bx}^s_i, y^s_i) \}_{i=1}^{n_s}$ and the reconstruction pipeline $f_\gen$ observes $\{ (\tilde{\bx}^t_i, \bx^t_i) \}_{i=1}^{n_t}$.
The noise distribution $\bbQ_{\tilde{X} | X}$ are typically geometric transformations (translation, rotation, skewing, and scaling) in data augmentation, while either zero-masked noise or Gaussian noise is used in the denoising strategy.
In this work, we combine all the fore-mentioned types of noise for denoising and use only the geometric transformations for data augmentation.

\section{Experiments and Results}
This section reports the evaluation results of $\mrcn$. It is divided into two parts.
The first part focuses on the evaluation on large-scale datasets popular with deep learning methods, 
while the second part summarizes the results on the Office dataset \cite{Saenko:2010aa}.

\subsection{Experiment I: SVHN, MNIST, USPS, CIFAR, and STL}
The first set of experiments investigates the empirical performance of $\mrcn$ on five widely used benchmarks: 
MNIST~\cite{LeCun:1998aa}, 
USPS~\cite{usps1994}, 
Street View House Numbers (SVHN)~\cite{svhn:2011},
CIFAR~\cite{Krizhevsky:2009aa},
and STL~\cite{Coates:2011ab}, see the corresponding references for more detailed configurations.
The task is to perform cross-domain recognition: \emph{taking the training set from one dataset as the source domain and the test set from another dataset as the target domain}. 
We evaluate our algorithm's recognition accuracy over three cross-domain pairs:
1) MNIST vs USPS, 2) SVHN vs MNIST, and 3) CIFAR vs STL.

MNIST (\textsc{mn}) vs USPS (\textsc{us}) contains 2D grayscale handwritten digit images of 10 classes.
We preprocessed them as follows. 
USPS images were rescaled into $28 \times 28$ and pixels were normalized to $[0,1]$ values.
From this pair, two cross-domain recognition tasks were performed:  \textsc{mn} $\rightarrow$  \textsc{us} and \textsc{us} $\rightarrow$ \textsc{mn}.

In SVHN (\textsc{sv}) vs MNIST (\textsc{mn}) pair, MNIST images were rescaled to $32 \times 32$ and SVHN images were grayscaled.
The $[0,1]$ normalization was then applied to all images. 
Note that we did not preprocess SVHN images using local contrast normalization as in \cite{Sermanet:2012}.
We evaluated our algorithm on \textsc{sv} $\rightarrow$ \textsc{mn} and \textsc{mn} $\rightarrow$ \textsc{sv} cross-domain recognition tasks.

STL (\textsc{st}) vs CIFAR (\textsc{ci}) consists of RGB images that share eight object classes: \emph{airplane}, \emph{bird}, \emph{cat}, \emph{deer}, \emph{dog}, \emph{horse}, \emph{ship}, and \emph{truck}, which forms $4,000$ (train) and $6,400$ (test) images for STL, and $40,000$ (train) and $8,000$ (test) images for CIFAR.
STL images were rescaled to $32 \times 32$ and pixels were standardized into zero-mean and unit-variance.
Our algorithm was evaluated on two cross-domain tasks, that is,  \textsc{st} $\rightarrow$ \textsc{ci} and \textsc{ci} $\rightarrow$ \textsc{st}.

\paragraph{\textbf{The architecture and learning setup:}}
The $\mrcn$ architecture used in the experiments is adopted from \cite{Masci2011}.
The label prediction pipeline has three convolutional layers: 100 5x5 filters (\textsc{conv1}), 150 5x5 filters (\textsc{conv2}), and 200 3x3 filters (\textsc{conv3}) respectively, two max-pooling layers of size 2x2 after the first and the second convolutional layers (\textsc{pool1} and \textsc{pool2}), and three fully-connected layers (\textsc{fc4}, \textsc{fc5},and \textsc{fc$\_$out}) -- \textsc{fc$\_$out} is the output layer.
The number of neurons in \textsc{fc4} or \textsc{fc5} was treated as a tunable hyper-parameter in the range of $[300, 350, ..., 1000]$, chosen according to the best performance on the validation set.
The shared encoder $g_\enc$ has thus a configuration of \textsc{conv1}-\textsc{pool1}-\textsc{conv2}-\textsc{pool2}-\textsc{conv3}-\textsc{fc4}-\textsc{fc5}.
Furthermore, the configuration of the decoder $g_\dec$ is the inverse of that of $g_\enc$.
Note that the unpooling operation in $g_\dec$ performs by upsampling-by-duplication: inserting the pooled values in the appropriate locations in the feature maps, with the remaining elements being the same as the pooled values.

We employ ReLU activations~\cite{Nair:2010aa} in all hidden layers and linear activations in the output layer of the reconstruction pipeline.
Updates in both classification and reconstruction tasks were computed via RMSprop with learning rate of $10^{-4}$ and moving average decay of $0.9$.
The control penalty $\lambda$ was selected according to accuracy on the source validation data -- typically, the optimal value was in the range $[0.4,0.7]$.

\paragraph{\textbf{Benchmark algorithms:}}
We compare DRCN with the following methods.
1) $\convnetsrc$: a supervised convolutional network trained on the labeled source domain only, with the same network configuration as that of $\mrcn$'s label prediction pipeline,
2) SCAE: ConvNet preceded by the layer-wise pretraining of stacked convolutional autoencoders on all unlabeled data~\cite{Masci2011},
3) $\scaet$: similar to SCAE, but only unlabeled data from the target domain are used during pretraining,
4) $\sda$~\cite{Glorot:2011aa}: the deep network with three fully connected layers, which is a successful domain adaptation model for sentiment classification,
5) Subspace Alignment (SA)~\cite{Fernando:2013aa},\footnote{The setup follows one in \cite{Ganin2015}: the inputs to SA are the last hidden layer activation values of $\convnetsrc$.} and
6) ReverseGrad~\cite{Ganin2015}: a recently published domain adaptation model based on deep convolutional networks that provides the state-of-the-art performance.

All deep learning based models above have the same architecture as DRCN for the label predictor.
For ReverseGrad, we also evaluated the ``original architecture'' devised in \cite{Ganin2015} and chose whichever performed better of the original architecture or our architecture.
Finally, we applied the data augmentation to all models similarly to $\mrcn$.
The ground-truth model is also evaluated, that is, a convolutional network trained from and tested on images from the target domain only ($\convnettgt$), to measure the difference between the cross-domain performance and the ideal performance.

\paragraph{\textbf{Classification accuracy:}}
Table~\ref{tab:acc} summarizes the cross-domain recognition accuracy (\emph{mean $\pm$ std}) of all algorithms over ten independent runs.
$\mrcn$ performs best in all but one cross-domain tasks, better than the prior state-of-the-art ReverseGrad.
Notably on the \textsc{sv} $\rightarrow$ \textsc{mn} task, $\mrcn$ outperforms ReverseGrad with $\sim8\%$ accuracy gap.
$\mrcn$ also provides a considerable improvement over ReverseGrad ($\sim5\%$) on the reverse task, \textsc{mn} $\rightarrow$ \textsc{sv}, 
but the gap to the groundtruth is still large -- this case was also mentioned in previous work as a failed case \cite{Ganin2015}.
In the case of \textsc{ci} $\rightarrow$ \textsc{st}, the performance of $\mrcn$ almost matches the performance of the target baseline.

$\mrcn$ also convincingly outperforms the greedy-layer pretraining-based algorithms ($\sda$, SCAE, and $\scaet$).
This indicates the effectiveness of the simultaneous reconstruction-classification training strategy over the standard pretraining-finetuning in the context of domain adaptation.

\begin{table}
    \caption{Accuracy (\emph{mean} $\pm$ \emph{std} $\%$) on five cross-domain recognition tasks over ten independent runs.
    Bold and underline indicate the best and second best domain adaptation performance.
    $\convnettgt$ denotes the ground-truth model: training and testing on the target domain only.
}
	\label{tab:acc}
\scalebox{0.80}{
\begin{tabular}{| l || c | c || c | c || c | c |}
\hline
Methods & $\textsc{mn} \rightarrow \textsc{us}$ & $\textsc{us} \rightarrow \textsc{mn}$ & $\textsc{sv} \rightarrow \textsc{mn}$ & $\textsc{mn} \rightarrow \textsc{sv}$ & $\textsc{st} \rightarrow \textsc{ci}$ & $\textsc{ci} \rightarrow \textsc{st}$ \\
\hline
$\convnetsrc$  & 85.55 $\pm$ 0.12 & 65.77 $\pm$ 0.06 & 62.33 $\pm$ 0.09 & 25.95 $\pm$ 0.04&  54.17 $\pm$ 0.21 & 63.61 $\pm$ 0.17 \\
\hline
$\sda$ \cite{Glorot:2011aa}& 43.14 $\pm$ 0.16 & 37.30 $\pm$ 0.12 & 55.15 $\pm$ 0.08 & 8.23 $\pm$ 0.11 & 35.82 $\pm$ 0.07 & 42.27 $\pm$ 0.12\\
SA \cite{Fernando:2013aa} & 85.89 $\pm$ 0.13 &  51.54 $\pm$ 0.06 & 63.17 $\pm$ 0.07 & 28.52 $\pm$ 0.10 & 54.04 $\pm$ 0.19 & 62.88 $\pm$ 0.15\\
SCAE \cite{Masci2011}  & 85.78 $\pm$ 0.08 & 63.11 $\pm$ 0.04 & 60.02 $\pm$ 0.16 & 27.12 $\pm$ 0.08& 54.25 $\pm$ 0.13 & 62.18 $\pm$ 0.04\\
$\scaet$ \cite{Masci2011}  & 86.24 $\pm$ 0.11  & 65.37 $\pm$ 0.03  & $65.57 \pm 0.09$ & 27.57 $\pm$ 0.13 &  $54.68 \pm 0.08$  & $61.94 \pm 0.06$ \\
ReverseGrad \cite{Ganin2015} & \underline{91.11 $\pm$ 0.07}  & \textbf{74.01 $\pm$ 0.05} & \underline{73.91 $\pm$ 0.07} & \underline{35.67 $\pm$ 0.04} & \underline{56.91 $\pm$ 0.05}  & \underline{66.12 $\pm$ 0.08}\\
$\bmrcn$ & \textbf{91.80 $\pm$ 0.09} & \underline{73.67 $\pm$ 0.04} & \textbf{81.97 $\pm$ 0.16} &  \textbf{40.05 $\pm$ 0.07} &  $\mathbf{58.86 \pm 0.07}$ & $\mathbf{66.37 \pm 0.10}$\\
\hline
$\convnettgt$ & $96.12 \pm 0.07$ & $98.67 \pm 0.04$ & $98.67 \pm 0.04$ & 91.52 $\pm$ 0.05 & $78.81 \pm 0.11$ & $66.50 \pm 0.07$\\
\hline
\end{tabular}
}
\end{table}

\paragraph{\textbf{Comparison of different $\bmrcn$ flavors:}}
Recall that $\mrcn$ uses only the unlabeled target images for the unsupervised reconstruction training.
To verify the importance of this strategy, we further compare different flavors of $\mrcn$: $\mrcns$ and $\mrcnst$.
Those algorithms are conceptually the same but different only in utilizing the unlabeled images during the unsupervised training.
$\mrcns$ uses only unlabeled source images, whereas $\mrcnst$ combines both unlabeled source and target images.

The experimental results in Table \ref{tab:acc2} confirm that $\mrcn$ always performs better than $\mrcns$ and $\mrcnst$.
While $\mrcnst$ occasionally outperforms ReverseGrad, its overall performance does not compete with that of $\mrcn$.
The only case where $\mrcns$ and $\mrcnst$ flavors can closely match $\mrcn$ is on \textsc{mn}$\rightarrow$ \textsc{us}.
This suggests that the use of \emph{unlabeled source data} during the reconstruction training do not contribute much to the cross-domain generalization, 
which verifies the $\mrcn$ strategy in using the unlabeled target data only.

\begin{table}
\caption{Accuracy ($\%$) of $\mrcns$ and $\mrcnst$.}
\label{tab:acc2}
\centering
\scalebox{0.85}{
\begin{tabular}{| l || c | c || c | c || c | c |}
\hline
Methods & $\textsc{mn} \rightarrow \textsc{us}$ & $\textsc{us} \rightarrow \textsc{mn}$ & $\textsc{sv} \rightarrow \textsc{mn}$ & $\textsc{mn} \rightarrow \textsc{sv}$ & $\textsc{st} \rightarrow \textsc{ci}$ & $\textsc{ci} \rightarrow \textsc{st}$ \\
\hline
$\bmrcns$ & 89.92 $\pm$ 0.12 & 65.96 $\pm$ 0.07 & 73.66 $\pm$ 0.04 & 34.29 $\pm$ 0.09 & 55.12 $\pm$ 0.12 & 63.02 $\pm$ 0.06\\
$\bmrcnst$  & 91.15 $\pm$ 0.05 & 68.64 $\pm$ 0.05 & 75.88 $\pm$ 0.09 & 37.77 $\pm$ 0.06 &  55.26 $\pm$ 0.06 & 64.55 $\pm$ 0.13\\
$\bmrcn$ & \textbf{91.80 $\pm$ 0.09} & \textbf{73.67 $\pm$ 0.04} & \textbf{81.97 $\pm$ 0.16} &  \textbf{40.05 $\pm$ 0.07} &  $\mathbf{58.86 \pm 0.07}$ & $\mathbf{66.37 \pm 0.10}$\\
\hline
\end{tabular}
}
\end{table}
    
\begin{figure}[!htb]
  \centering
    	\subfigure[Source (SVHN)]{\includegraphics[width=1.55in,height=1.45in]{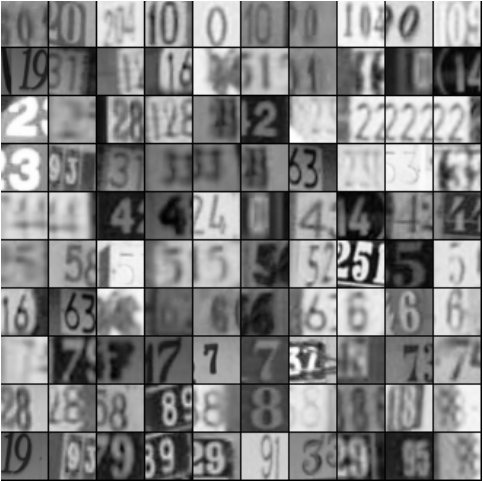} \label{fig:sv}}  \quad \quad \quad
	\subfigure[Target (MNIST) ]{\includegraphics[width=1.55in,height=1.45in]{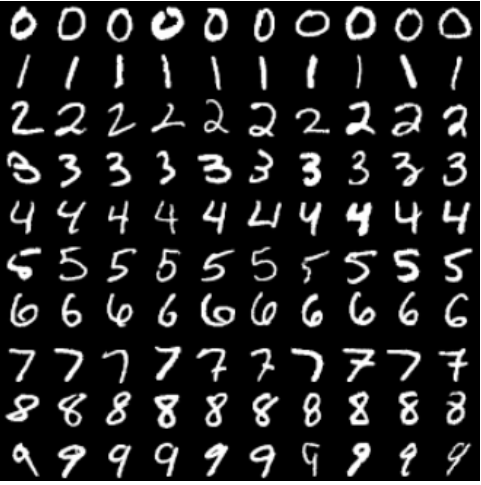} \label{fig:mi}} \quad \quad \quad
	\subfigure[$\mrcn$]{\includegraphics[width=1.55in,height=1.45in]{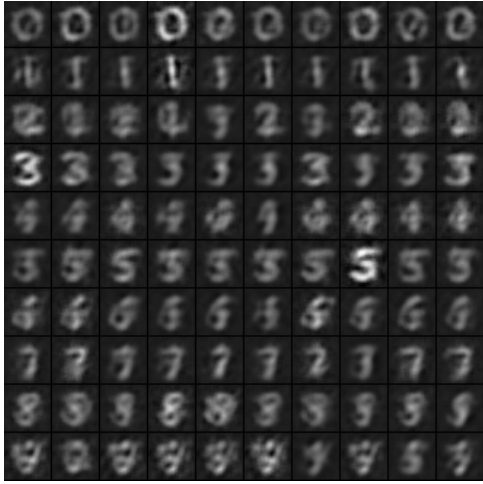} \label{fig:drcn_rec}}  \quad \quad \quad
    	\subfigure[ConvAE]{\includegraphics[width=1.55in,height=1.45in]{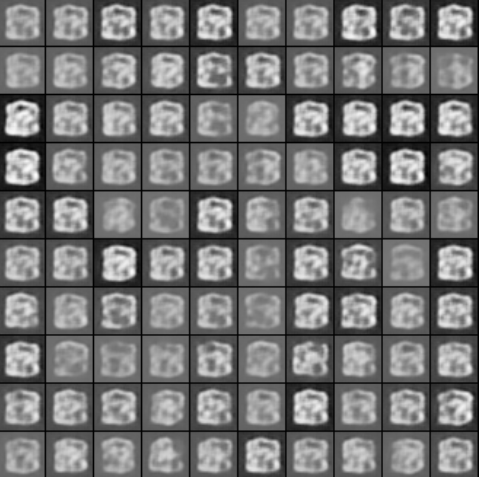} \label{fig:convae_rec}} \quad \quad \quad
	\subfigure[$\mrcnst$]{\includegraphics[width=1.55in,height=1.45in]{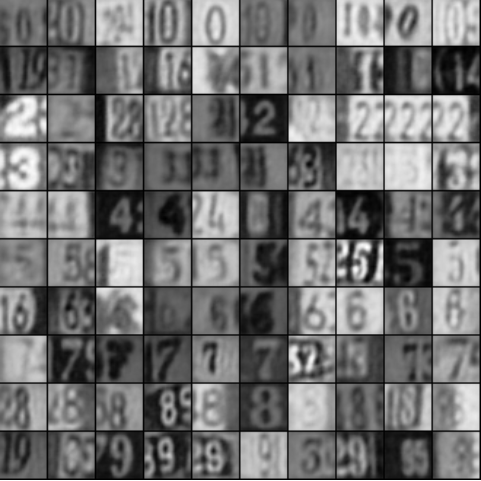}
	\label{fig:drcnst_rec}}  \quad \quad \quad
	\subfigure[ConvAE+ConvNet]{\includegraphics[width=1.55in,height=1.45in]{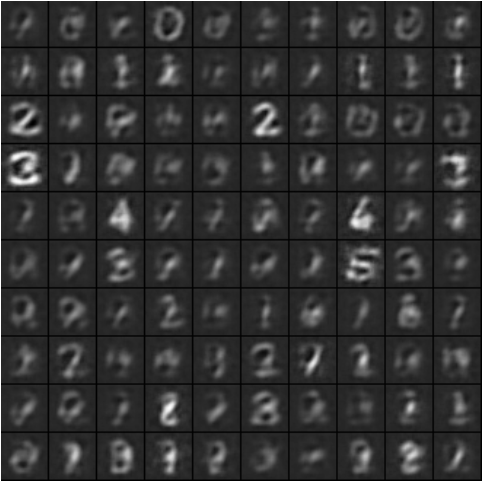} \label{fig:convae-convnet_rec}} 
	\caption{Data reconstruction after training from SVHN $\rightarrow$ MNIST. 
	Fig. (a)-(b) show the original input pixels, and (c)-(f) depict the reconstructed source images (SVHN).
	The reconstruction of $\mrcn$ appears to be MNIST-like digits, see the main text for a detailed explanation.}
	\label{fig:rec}
\end{figure}

\paragraph{\textbf{Data reconstruction:}}
A useful insight was found when reconstructing source images through the reconstruction pipeline of $\mrcn$. 
Specifically, we observe the visual appearance of $f_\gen(x^s_1), \ldots, f_\gen(x^s_m)$, where $x^s_1, \ldots, x^s_m$ are some images from the source domain.
Note that $x^s_1, \ldots, x^s_m$ are unseen during the unsupervised reconstruction training in $\mrcn$.
We visualize such a reconstruction in the case of \textsc{sv} $\rightarrow$\textsc{mn} training in Figure \ref{fig:rec}.
Figure \ref{fig:sv} and \ref{fig:mi} display the original source (SVHN) and target (MNIST) images. 

The main finding of this observation is depicted in Figure \ref{fig:drcn_rec}:
the reconstructed images produced by $\mrcn$ given some SVHN images as the source inputs.
We found that \emph{the reconstructed SVHN images resemble MNIST-like digit appearances, with white stroke and black background}, see Figure \ref{fig:mi}.
Remarkably, $\mrcn$ still can produce ``correct" reconstructions of some noisy SVHN images.
For example, all SVHN digits 3 displayed in Figure \ref{fig:sv} are clearly reconstructed by $\mrcn$, 
see the fourth row of Figure \ref{fig:drcn_rec}.
$\mrcn$ tends to pick only the digit in the middle and ignore the remaining digits.
This may explain the superior cross-domain recognition performance of $\mrcn$ on this task. 
However, such a cross-reconstruction appearance does not happen in the reverse task, \textsc{mn} $\rightarrow$ \textsc{sv}, 
which may be an indicator for the low accuracy relative to the groundtruth performance.

We also conduct such a diagnostic reconstruction on other algorithms that have the reconstruction pipeline.
Figure \ref{fig:convae_rec} depicts the reconstructions of the SVHN images produced by ConvAE trained on the MNIST images only.
They do not appear to be digits, suggesting that ConvAE recognizes the SVHN images as noise.
Figure \ref{fig:drcnst_rec} shows the reconstructed SVHN images produced by $\mrcnst$.
We can see that they look almost identical to the source images shown in Figure \ref{fig:sv}, which is not surprising since the source images are included during the reconstruction training.

Finally, we evaluated the reconstruction induced by $\convnetsrc$ to observe the difference with the reconstruction of $\mrcn$.
Specifically, we trained ConvAE on the MNIST images in which the encoding parameters were initialized from those of $\convnetsrc$ and not updated during training.
We refer to the model as ConvAE+$\convnetsrc$.
The reconstructed images are visualized in Figure \ref{fig:convae-convnet_rec}.
Although they resemble the style of MNIST images as in the $\mrcn$'s case, only a few source images are correctly reconstructed.

To summarize, the results from this diagnostic data reconstruction correlate with the cross-domain recognition performance.
More visualization on other cross-domain cases can be found in the Supplemental materials.

\subsection{Experiments II: Office dataset}
In the second experiment, we evaluated $\mrcn$ on the standard domain adaptation benchmark for visual object recognition, \textsc{Office} \cite{Saenko:2010aa}, which consists of three different domains: \textsc{amazon (a)}, \textsc{dslr (d)}, and \textsc{webcam (w)}.
\textsc{Office} has 2817 labeled images in total distributed across 31 object categories. 
The number of images is thus relatively small compared to the previously used datasets.

We applied the $\mrcn$ algorithm to \emph{finetune} AlexNet \cite{Krizhevsky_NIPS2012}, as was done with different methods in previous work \cite{Ganin2015,Long_DAN:2015,Tzeng_DDC:2014}.\footnote{Recall that AlexNet consists of five convolutional layers: $conv1, \ldots, conv5$ and three fully connected layers: $fc6, fc7$, and $fc8/output$.}
The fine-tuning was performed only on the fully connected layers of AlexNet, $fc6$ and $fc7$, and the last convolutional layer, $conv5$.
Specifically, the label prediction pipeline of $\mrcn$ contains  $conv4$-$conv5$-$fc6$-$fc7$-$label$ and the data reconstruction pipeline has $conv4$-$conv5$-$fc6$-$fc7$-$fc6'$-$conv5'$-$conv4'$ (the $'$ denotes the the inverse layer) -- it thus does not reconstruct the original input pixels.
The learning rate was selected following the strategy devised in \cite{Long_DAN:2015}: cross-validating the base learning rate between $10^{-5}$ and $10^{-2}$ with a multiplicative step-size $10^{1/2}$.

We followed the standard unsupervised domain adaptation training protocol used in previous work \cite{Chopra:2013aa,Gong:2013ab,Long_DAN:2015}, that is, using \emph{all} labeled source data and unlabeled target data.
Table \ref{tab:office_acc} summarizes the performance accuracy of $\mrcn$ based on that protocol in comparison to the state-of-the-art algorithms.
We found that $\mrcn$ is competitive against DAN and ReverseGrad -- the performance is either the best or the second best except for one case.
In particular, $\mrcn$ performs best with a convincing gap in situations when the target domain has relatively many data, i.e., $\textsc{amazon}$ as the target dataset.

\begin{table}[!htb]
\centering
\caption{Accuracy (\emph{mean} $\pm$ \emph{std} $\%$) on the Office dataset with the standard unsupervised domain adaptation protocol used in \cite{Gong:2013ab,Chopra:2013aa}.}
\label{tab:office_acc}
\scalebox{0.93}{
\begin{tabular}{| l | c | c | c | c | c | c |}
\hline
Method & \textsc{a} $\rightarrow$ \textsc{w} & \textsc{w} $\rightarrow$ \textsc{a} & \textsc{a} $\rightarrow$ \textsc{d} & \textsc{d} $\rightarrow$ \textsc{a} & \textsc{w} $\rightarrow$ \textsc{d} & \textsc{d} $\rightarrow$ \textsc{w}\\
\hline
DDC \cite{Tzeng_DDC:2014} & 61.8 $\pm$ 0.4 & 52.2 $\pm$ 0.4 & 64.4 $\pm$ 0.3 & 52.1 $\pm$ 0.8 & 98.5 $\pm$ 0.4 & 95.0 $\pm$ 0.5 \\
DAN \cite{Long_DAN:2015} & 68.5 $\pm$ 0.4 & \underline{53.1} $\pm$ 0.3 & \underline{67.0} $\pm$ 0.4 & 54.0 $\pm$ 0.4 & \underline{99.0} $\pm$ 0.2 & \underline{96.0} $\pm$ 0.3 \\
ReverseGrad \cite{Ganin2015} & \textbf{72.6} $\pm$ 0.3 & 52.7 $\pm$ 0.2 & \textbf{67.1} $\pm$ 0.3 & \underline{54.5} $\pm$ 0.4 & \textbf{99.2} $\pm$ 0.3& \textbf{96.4} $\pm$ 0.1\\
$\bmrcn$ & \underline{68.7} $\pm$ 0.3& \textbf{54.9} $\pm$ 0.5 & 66.8 $\pm$ 0.5 & \textbf{56.0} $\pm$ 0.5 & \underline{99.0} $\pm$ 0.2& \textbf{96.4} $\pm$ 0.3 \\
\hline
\end{tabular}
}
\end{table}

\section{Analysis} 
\label{sec:analysis}
This section provides a first step towards a formal analysis of the DRCN algorithm.
We demonstrate that optimizing \eqref{eq:mdgn_obj} in $\mrcn$ relates to solving a semi-supervised learning problem on the target domain according to a framework proposed in \cite{Cohen:2006}.
The analysis suggests that unsupervised training using only unlabeled target data is sufficient.
That is, adding unlabeled source data might not further improve domain adaptation.

Denote the labeled and unlabeled distributions as $\bbD_{XY} =: \bbD$ and $\bbD_{X}$ respectively. Let $P^\theta(\cdot)$ refer to a family of models, parameterized by $\theta\in\Theta$, that is used to learn a maximum likelihood estimator. The $\mrcn$ learning algorithm for domain adaptation tasks can be interpreted probabilistically by assuming that $P^\theta(x)$ is Gaussian and $P^\theta(y|x)$ is a multinomial distribution, fit by logistic regression. 

The objective in Eq.\eqref{eq:mdgn_obj} is  equivalent to the following maximum likelihood estimate:
\begin{equation}
  \label{eq:approx_target}
  \hat{\theta} =
  \argmax_{\theta} \lambda \sum_{i=1}^{n_s} \log P^{\theta}_{Y|X} (y^s_i | x^s_i)  + (1 - \lambda) \sum_{j=1}^{n_t} \log P^{\theta}_{X | \tilde{X}}(x^t_j | \tilde{x}^t_j),
\end{equation}
where $\tilde{x}$ is the noisy input generated from $\bbQ_{\tilde{X} | X }$.
The first term represents the model learned by the supervised convolutional network and the second term represents the model learned by the unsupervised convolutional autoencoder.
Note that the discriminative model only observes labeled data from the source distribution $\bbP_X$ in objectives \eqref{eq:mdgn_obj} and \eqref{eq:approx_target}.

We now recall a semi-supervised learning problem formulated in \cite{Cohen:2006}. 
Suppose that labeled and unlabeled samples are taken from the \emph{target domain} $\bbQ$ with probabilities $\lambda$ and $(1-\lambda)$ respectively.
By Theorem 5.1 in \cite{Cohen:2006}, the maximum likelihood estimate $\zeta$ is
  \begin{equation}
    \label{eq:true_target}
	\zeta = \argmax_{\zeta} \lambda \expec_{\bbQ} [\log P^{\zeta}(x, y)]+ (1 - \lambda) \expec_{\bbQ_X }[\log P^{\zeta}_X(x)]
  \end{equation}
The theorem holds if it satisfies the following assumptions:
\emph{consistency}, the model contains true distribution, so the MLE is consistent; and 
\emph{smoothness and measurability} \cite{White:1982}.
Given target data $ (x_1^t, y_1^t) , \ldots, (x_{n_t}^t, y_{n_t}^t) \sim \bbQ$, the parameter $\zeta$ can be estimated as follows:
  \begin{equation}
    \label{eq:true_target2}
	\hat{\zeta} = \argmax_{\zeta} \lambda \sum_{i=1}^{n_t} [\log P^{\zeta}(x_i^t, y_i^t)] + (1 - \lambda) \sum_{i=1}^{n_t}[\log P^{\zeta}_X(x_i^t)]
  \end{equation}
Unfortunately, $\hat{\zeta}$ cannot be computed in the unsupervised domain adaptation setting since we do not have access to target labels. 

Next we inspect a certain condition where $\hat{\theta}$ and $\hat{\zeta}$ are closely related.
Firstly, by the \emph{covariate shift} assumption \cite{Shimodaira:2000aa}: $\bbP \neq \bbQ$ and $\bbP_{Y|X} = \bbQ_{Y | X}$, the first term in (\ref{eq:true_target}) can be switched from an expectation over target samples to source samples:
  \begin{equation}
    \expec_{\bbQ} \Big[\log P^{\zeta}(x, y)\Big] = \expec_{\bbP}\left[\frac{\bbQ_X(x)}{\bbP_X(x)}\cdot \log P^{\zeta}(x, y)\right].
  \end{equation}
Secondly, it was shown in \cite{Bengio_gdae:2013} that $P^\theta_{X | \tilde{X}}(x | \tilde{x})$, see the second term in (\ref{eq:approx_target}),  defines an ergodic Markov chain whose asymptotic marginal distribution of $X$ converges to the data-generating distribution $\bbP_X$.
Hence, Eq. (\ref{eq:true_target2}) can be rewritten as
  \begin{equation}
    \label{eq:true_target3}
	\hat{\zeta} \approx \argmax_{\zeta} \lambda \sum_{i=1}^{n_s} \frac{\bbQ_X(x_i^s)}{\bbP_X(x_i^s)} \log P^\zeta(x_i^s, y_i^s) + (1 - \lambda) \sum_{j=1}^{n_t}[\log P^{\zeta}_{X|\tilde{X}}(x_j^t | \tilde{x}_j^t)].
  \end{equation}
The above objective differs from objective (\ref{eq:approx_target}) only in the first term.
Notice that $\hat{\zeta}$ would be approximately equal $\hat{\theta}$ if the ratio $\frac{\bbQ_X(x_i^s)}{\bbP_X(x_i^s)}$ is constant for all $x^s$. 
In fact, it becomes the objective of $\mrcnst$.
Although the constant ratio assumption is too strong to hold in practice, comparing \eqref{eq:approx_target} and \eqref{eq:true_target3} suggests that $\hat{\zeta}$ can be a reasonable approximation to $\hat{\theta}$.

Finally, we argue that using unlabeled source samples during the unsupervised training may not further contribute to domain adaptation.
To see this, we expand the first term of (\ref{eq:true_target3}) as follows
  \begin{equation}
  \lambda \sum_{i=1}^{n_s} \frac{\bbQ_X(x^s_i)}{\bbP_X(x^s_i)}  \log P^\zeta_{Y|X}(y^s_i | x^s_i) + 
  \lambda \sum_{i=1}^{n_s} \frac{\bbQ_X(x^s_i)}{\bbP_X(x^s_i)} \log P^\zeta_X(x^s_i). \nonumber
  \end{equation}
Observe the second term above. As $n_s \rightarrow \infty$, $P^\theta_X$ will converge to $\bbP_X$.
Hence, since
$\int_{x \sim \bbP_X} \frac{\bbQ_X(x)}{\bbP_X(x)} \log \bbP_X(x) \leq \int_{x \sim \bbP_X} \bbP_X^t(x)$, adding more unlabeled source data will only result in a constant.
This implies an optimization procedure equivalent to  (\ref{eq:approx_target}), which may explain the \emph{uselessness} of unlabeled source data in the context of domain adaptation.

Note that the latter analysis does not necessarily imply that incorporating unlabeled source data degrades the performance.
The fact that $\mrcnst$ performs worse than $\mrcn$ could be due to, e.g., the model capacity, which depends on the choice of the architecture.

\section{Conclusions} 
We have proposed Deep Reconstruction-Classification Network ($\mrcn$), a novel model for unsupervised domain adaptation in object recognition.
The model performs multitask learning, i.e., alternately learning (source) label prediction and (target) data reconstruction using a shared encoding representation.
We have shown that $\mrcn$ provides a considerable improvement for some cross-domain recognition tasks over the state-of-the-art model.
It also performs better than deep models trained using the standard \emph{pretraining-finetuning} approach.
A useful insight into the effectiveness of the learned $\mrcn$ can be obtained from its data reconstruction.
The appearance of $\mrcn$'s reconstructed source images resemble that of the target images, which indicates that $\mrcn$ learns the domain correspondence.
We also provided a theoretical analysis relating the $\mrcn$ algorithm to semi-supervised learning. 
The analysis was used to support the strategy in involving only the target unlabeled data during learning the reconstruction task.

\clearpage

\bibliographystyle{splncs}
\bibliography{../mdgn}

\clearpage

\begin{center}
\Large{\textbf{Supplemental Material}} 
\end{center}

\begin{sloppypar}
This document is the supplemental material for the paper \emph{Deep Reconstruction-Classification for Unsupervised Domain Adaptation}.
It contains some more experimental results that cannot be included in the main manuscript due to a lack of space.
\end{sloppypar}

\begin{figure}[!htb]
  \centering
    	\subfigure[Source (MNIST)]{\includegraphics[width=1.55in,height=1.45in]{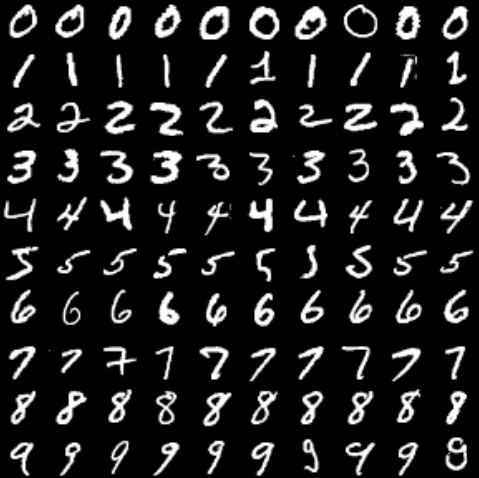} \label{fig:mi}}  \quad \quad \quad
	\subfigure[Target (USPS)]{\includegraphics[width=1.55in,height=1.45in]{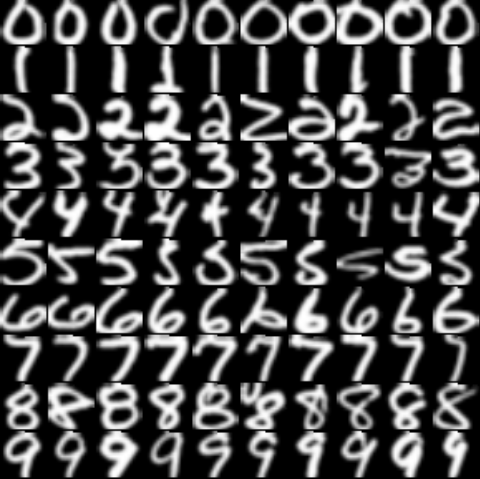} \label{fig:us}} \quad \quad \quad
	\subfigure[$\mrcn$]{\includegraphics[width=1.55in,height=1.45in]{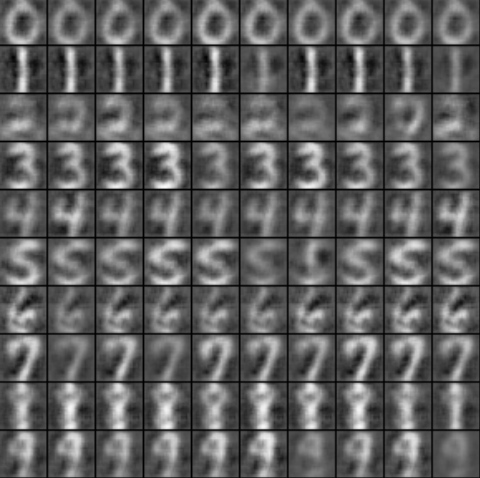} \label{fig:drcn_rec}}  \quad \quad \quad
    	\subfigure[ConvAE]{\includegraphics[width=1.55in,height=1.45in]{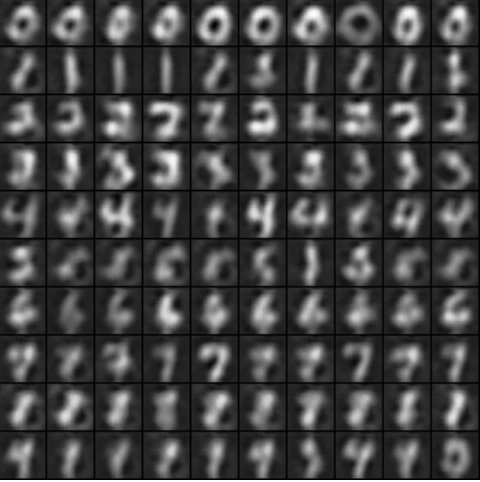} \label{fig:convae_rec}} \quad \quad \quad
	\subfigure[$\mrcnst$]{\includegraphics[width=1.55in,height=1.45in]{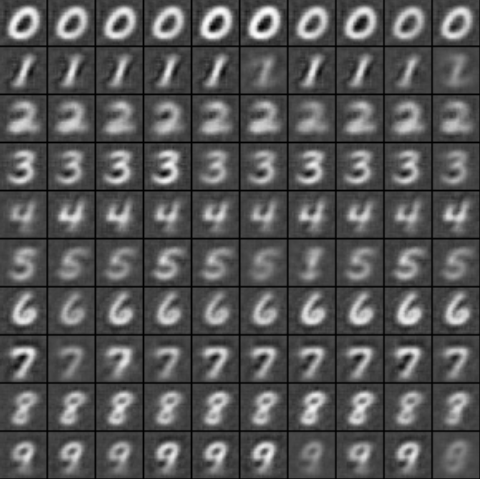}
	\label{fig:drcnst_rec}}  \quad \quad \quad
	\subfigure[ConvAE+$\textnormal{ConvNet}_{src}$]{\includegraphics[width=1.55in,height=1.45in]{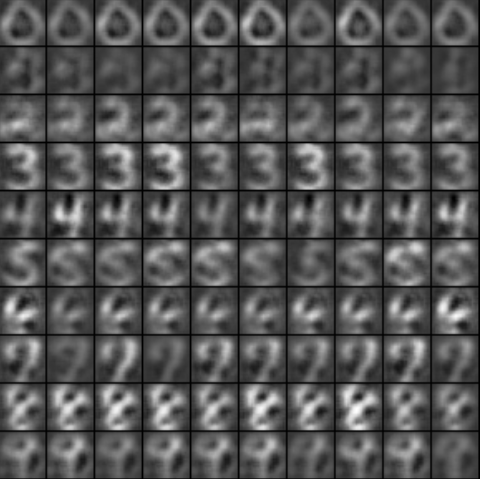} \label{fig:convae-convnet_rec}} 
	\vspace{-1em}
	\caption{Data reconstruction after training from MNIST $\rightarrow$ USPS. 
	Fig. (a)-(b) show the original input pixels, and (c)-(f) depict the reconstructed source images (MNIST).
}
	\label{fig:rec}
\end{figure}

\begin{figure}[!htb]
  \centering
    	\subfigure[Source (USPS)]{\includegraphics[width=1.55in,height=1.45in]{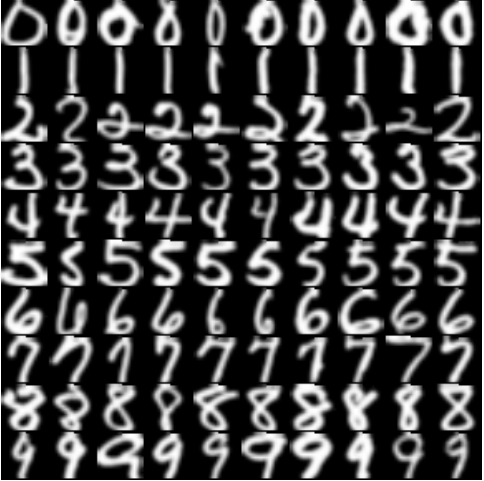} \label{fig:us2}}  \quad \quad \quad
	\subfigure[Target (MNIST)]{\includegraphics[width=1.55in,height=1.45in]{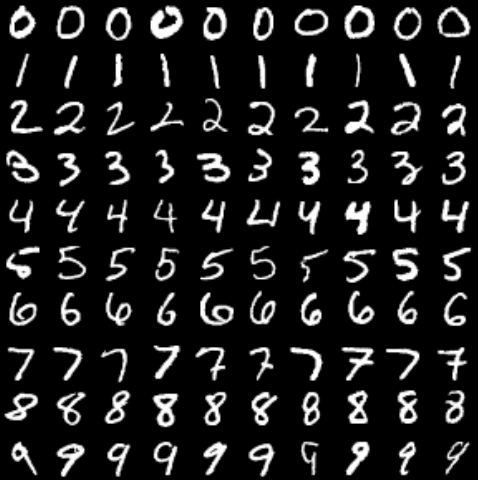} \label{fig:mi2}} \quad \quad \quad
	\subfigure[$\mrcn$]{\includegraphics[width=1.55in,height=1.45in]{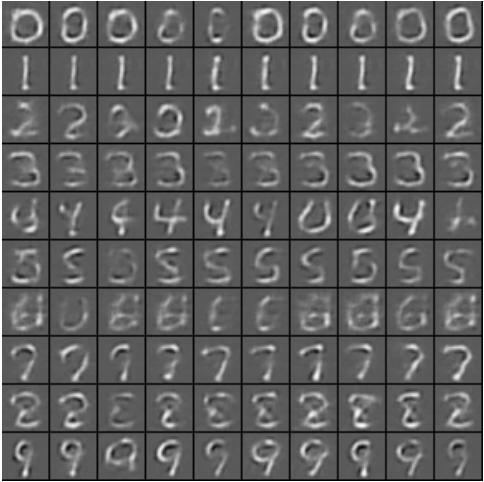} \label{fig:drcn_rec2}}  \quad \quad \quad
    	\subfigure[ConvAE]{\includegraphics[width=1.55in,height=1.45in]{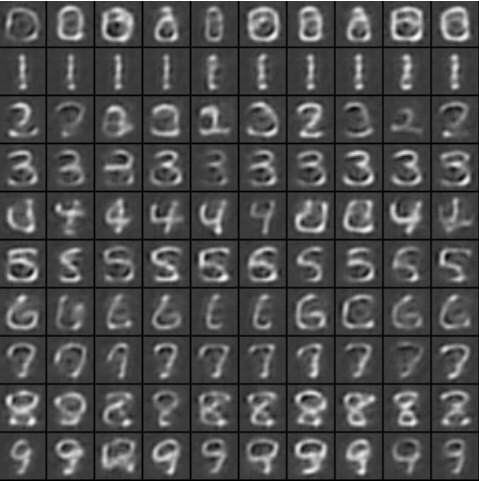} \label{fig:convae_rec2}} \quad \quad \quad
	\subfigure[$\mrcnst$]{\includegraphics[width=1.55in,height=1.45in]{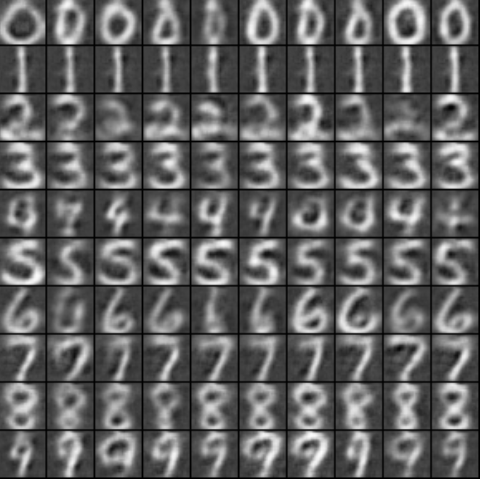}
	\label{fig:drcnst_rec2}}  \quad \quad \quad
	\subfigure[ConvAE+$\textnormal{ConvNet}_{src}$]{\includegraphics[width=1.55in,height=1.45in]{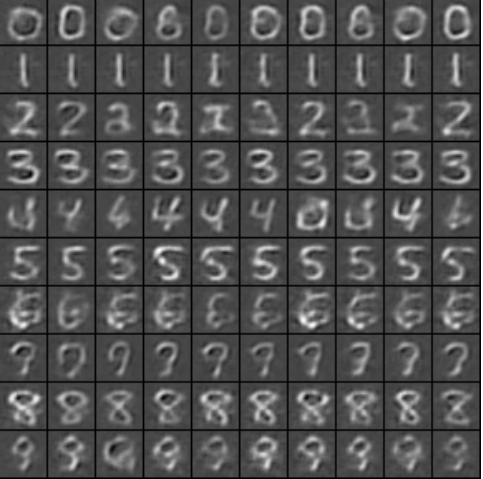} \label{fig:convae-convnet_rec2}}
	\caption{Data reconstruction after training from USPS $\rightarrow$ MNIST. 
	Fig. (a)-(b) show the original input pixels, and (c)-(f) depict the reconstructed source images (USPS).
}
	\label{fig:rec2}
\end{figure}
\vspace{-1em}
\section*{Data Reconstruction}
\vspace{-1em}
Figures \ref{fig:rec} and \ref{fig:rec2} depict the reconstruction of the source images in cases of MNIST $\rightarrow$ USPS and USPS $\rightarrow$ MNIST, respectively.
The trend of the outcome is similar to that of SVHN $\rightarrow$ MNIST, see Figure 2 in the main manuscript.
That is, the reconstructed images produced by $\mrcn$ resemble the \emph{style} of the target images.

\section*{Training Progress}
\begin{figure}[!htb]
 \centering
 \subfigure[SVHN $\rightarrow$ MNIST training]{\includegraphics[width=2.2in]{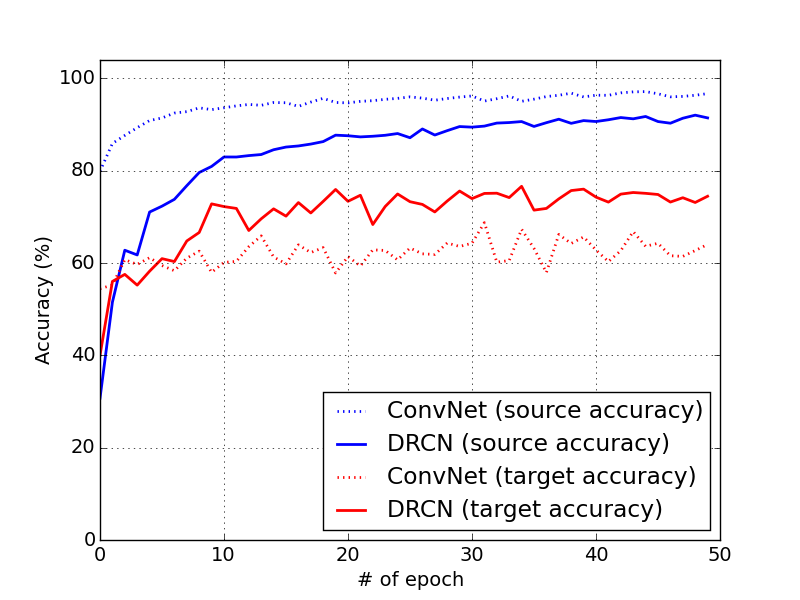}} \quad\quad
 \subfigure[MNIST $\rightarrow$ USPS training]{\includegraphics[width=2.2in]{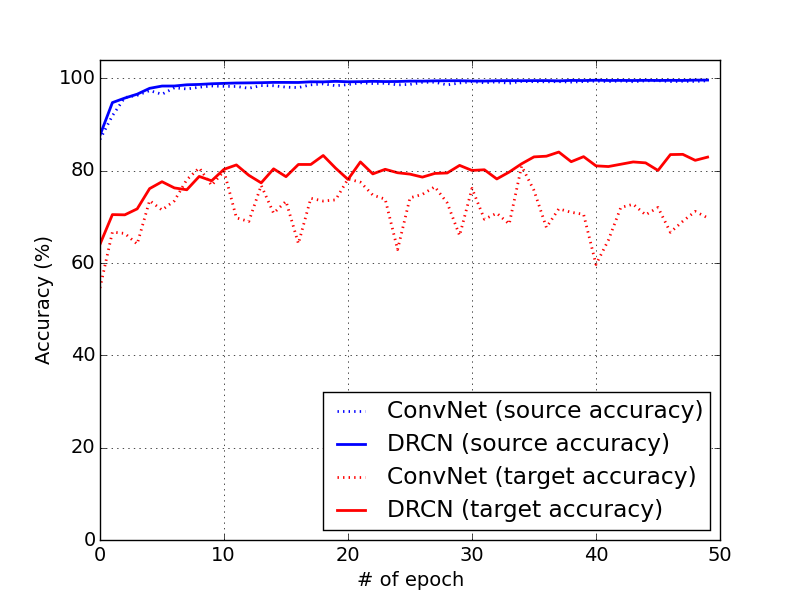}}
 \caption{The source accuracy (blue lines) and target accuracy (red lines) comparison between ConvNet and $\mrcn$ during training stage on SVHN $\rightarrow$ MNIST cross-domain task.
 $\mrcn$ induces lower source accuracy, but higher target accuracy than ConvNet.
 }
 \label{fig:accs_plot}
\end{figure}

Recall that $\mrcn$ has two pipelines with a shared encoding representation; each corresponds to the classification and reconstruction task, respectively.
One can consider that the unsupervised reconstruction learning acts as a regularization for the supervised classification to reduce overfitting onto the source domain.
Figure~\ref{fig:accs_plot} compares the source and target accuracy of $\mrcn$ with that of the standard ConvNet during training.
The most prominent results indicating the overfitting reduction can be seen in SVHN $\rightarrow$ MNIST case, i.e., $\mrcn$ produces higher target accuracy, but with lower source accuracy, than ConvNet.

\section*{t-SNE visualization.}
\begin{figure}[!htb]
  \centering
  \subfigure[ConvNet (MNIST $\rightarrow$ USPS)]{\includegraphics[width=1.62in]{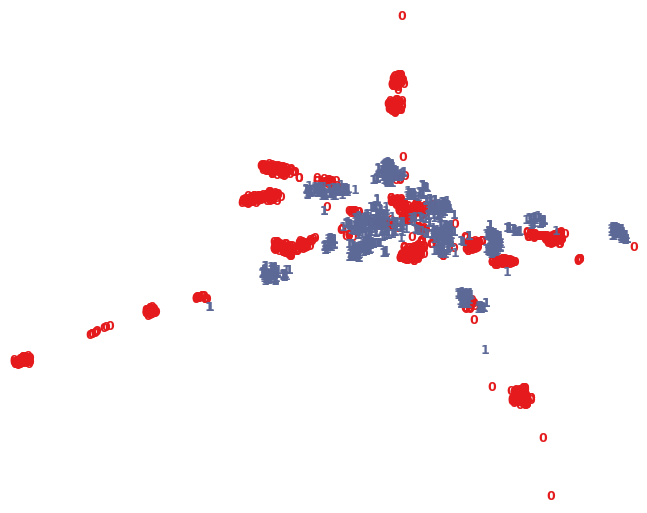}} \quad \quad \quad \quad
  \subfigure[$\mrcn$ (MNIST $\rightarrow$ USPS)]{\includegraphics[width=1.62in]{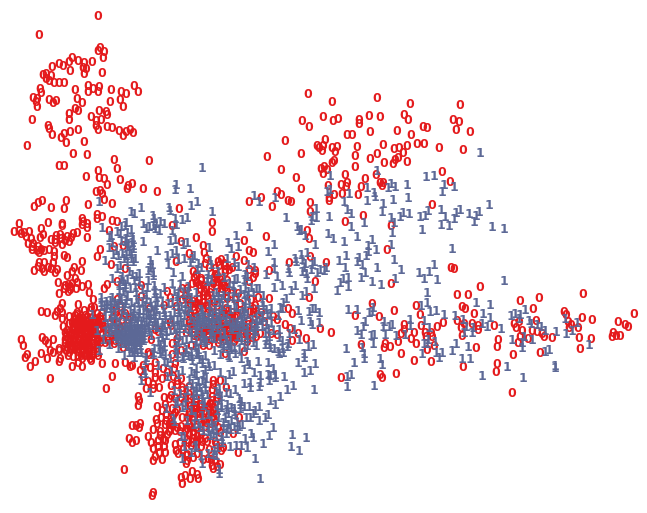}}
  \subfigure[ConvNet (SVHN $\rightarrow$ MNIST)]{\includegraphics[width=1.62in]{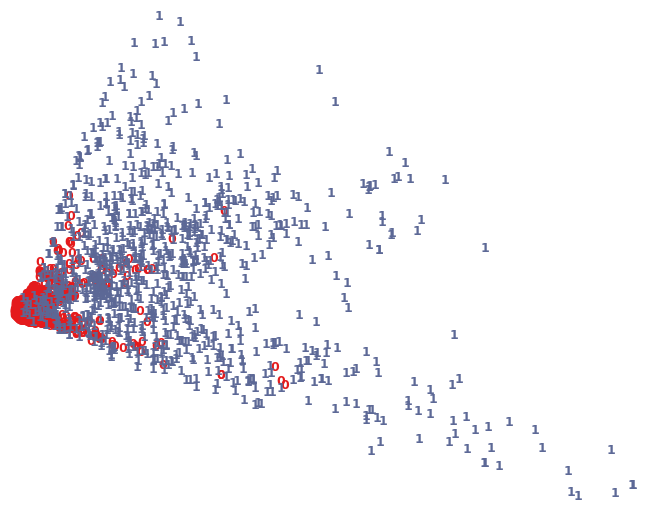}} \quad \quad \quad \quad
  \subfigure[$\mrcn$ (SVHN $\rightarrow$ MNIST)]{\includegraphics[width=1.62in]{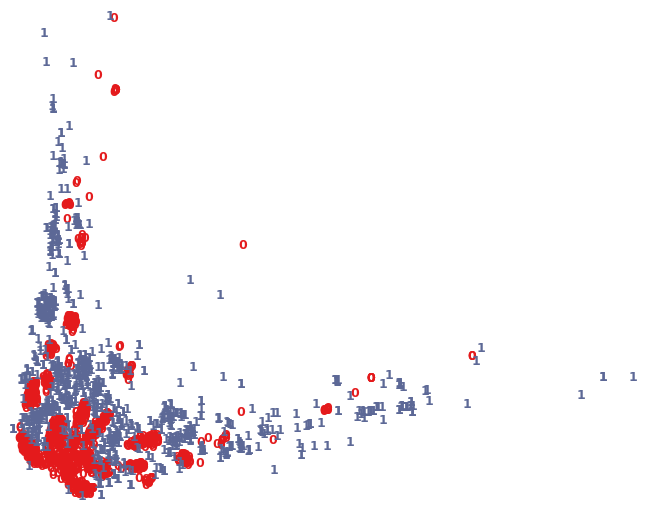}} 
  \caption{The t-SNE visualizations of the last layer's activations. Red and gray points indicate the source and target domain examples, respectively. 
  }
  \label{fig:tsne}
\end{figure}

For completeness, we also visualize the 2D point cloud of the last hidden layer of $\mrcn$ using t-SNE~\cite{Maaten:2008aa} and compare it with that of the standard ConvNet.
Figure \ref{fig:tsne} depicts the feature-point clouds extracted from the target images in the case of MNIST $\rightarrow$ USPS and SVHN $\rightarrow$ MNIST.
Red points indicate the source feature-point cloud, while gray points indicate the target feature-point cloud.
Domain invariance should be indicated by the degree of overlap between the source and target feature clouds.
We can see that the overlap is more prominent in the case of $\mrcn$ than $\convnet$.

\end{document}